\documentclass[11pt]{article}

\usepackage[final]{acl}

\usepackage{times}
\usepackage{latexsym}
\usepackage{booktabs}
\usepackage{multirow}
\usepackage{stfloats}

\usepackage[T1]{fontenc}
\usepackage[utf8]{inputenc}

\usepackage{microtype}

\usepackage{inconsolata}

\usepackage{graphicx}
\usepackage{siunitx}

\usepackage{comment}

\usepackage{listings}
\usepackage{xcolor}
\colorlet{punct}{red!60!black}
\definecolor{delim}{RGB}{20,105,176}
\colorlet{numb}{magenta!60!black}

\lstdefinelanguage{json}{
    basicstyle=\normalfont\ttfamily,
    numberstyle=\scriptsize,
    stepnumber=1,
    numbersep=8pt,
    showstringspaces=false,
    breaklines=true,
    literate=
     *{0}{{{\color{numb}0}}}{1}
      {1}{{{\color{numb}1}}}{1}
      {2}{{{\color{numb}2}}}{1}
      {3}{{{\color{numb}3}}}{1}
      {4}{{{\color{numb}4}}}{1}
      {5}{{{\color{numb}5}}}{1}
      {6}{{{\color{numb}6}}}{1}
      {7}{{{\color{numb}7}}}{1}
      {8}{{{\color{numb}8}}}{1}
      {9}{{{\color{numb}9}}}{1}
      {:}{{{\color{punct}{:}}}}{1}
      {,}{{{\color{punct}{,}}}}{1}
      {\{}{{{\color{delim}{\{}}}}{1}
      {\}}{{{\color{delim}{\}}}}}{1}
      {[}{{{\color{delim}{[}}}}{1}
      {]}{{{\color{delim}{]}}}}{1},
}

\title{\textsc{PartisanLens}: A Multilingual Dataset of Hyperpartisan and Conspiratorial Immigration Narratives in European Media}

\newcommand{\equalcontrib}{\textsuperscript{\textnormal{\fnsymbol{footnote}}}}

\author{  
  \textbf{Michele Joshua Maggini\textsuperscript{1}}\thanks{Michele Joshua Maggini and Paloma Piot contributed equally to this work.},
  \textbf{Paloma Piot\textsuperscript{2}}\equalcontrib,
  \textbf{Anxo Pérez\textsuperscript{2}}, \\
  \textbf{Erik Bran Marino\textsuperscript{3}},
  \textbf{Lúa Santamaría Montesinos\textsuperscript{4}},
  \textbf{Ana Lisboa\textsuperscript{5}}, \\
  \textbf{Marta Vázquez Abuín\textsuperscript{1}}, 
  \textbf{Javier Parapar \textsuperscript{2}},
  \textbf{Pablo Gamallo\textsuperscript{1}},
\\
  \textsuperscript{1}Centro Singular de Investigación en Tecnoloxías Intelixentes da USC, \\
  \textsuperscript{2}IRLab, CITIC Research Centre, Universidade da Coruña, 
  \textsuperscript{3}Universidade de Évora, \\
  \textsuperscript{4}Universidad de La Rioja,
  \textsuperscript{5}GESIS Leibniz Institute for the Social Sciences
\\
\\
  \small{
    \textbf{Correspondence:} \href{mailto:michelejoshua.maggini@usc.es}{michelejoshua.maggini@usc.es} and \href{mailto:paloma.piot@udc.es}{paloma.piot@udc.es}
  }
}

\begin{document}
\maketitle
\begin{abstract}

Detecting hyperpartisan narratives and Population Replacement Conspiracy Theories (PRCT) is essential to addressing the spread of misinformation. These complex narratives pose a significant threat, as hyperpartisanship drives political polarisation and institutional distrust, while PRCTs directly motivate real-world extremist violence, making their identification critical for social cohesion and public safety. However, existing resources are scarce, predominantly English-centric, and often analyse hyperpartisanship, stance, and rhetorical bias in isolation rather than as interrelated aspects of political discourse. To bridge this gap, we introduce \textsc{PartisanLens}, the first multilingual dataset of \num{1617} hyperpartisan news headlines in Spanish, Italian, and Portuguese, annotated in multiple political discourse aspects. We first evaluate the classification performance of widely used Large Language Models (LLMs) on this dataset, establishing robust baselines for the classification of hyperpartisan and PRCT narratives. In addition, we assess the viability of using LLMs as automatic annotators for this task, analysing their ability to approximate human annotation. Results highlight both their potential and current limitations. Next, moving beyond standard judgments, we explore whether LLMs can emulate human annotation patterns by conditioning them on socio-economic and ideological profiles that simulate annotator perspectives. At last, we provide our resources and evaluation, \textsc{PartisanLens} supports future research on detecting partisan and conspiratorial narratives in European contexts.
\end{abstract}

\section{Introduction}
\label{sec:Introduction}

The spread of misinformation threatens democratic societies by distorting public understanding and reducing trust in institutions. This issue is particularly serious in debates about immigration, where misleading stories can shape public opinion, amplify fear, and increase social divisions~\cite{marino2024polarization}. In Europe, this risk is growing as false or biased content influences migration discourses~\cite{soc13070168}. Two main harmful forms are hyperpartisan narratives and Population Replacement Conspiracy Theories (PRCTs), which fuel hostility toward migrants~\cite{dochow2024opinion}. Hyperpartisan content promotes extreme views through one-sided, emotional language~\cite{kiesel-etal-2019-semeval,systematichp}, while PRCTs spread the false idea that native populations are being deliberately replaced by migrants~\cite{marino2024polarization, sedgwick2024great}, reinforcing fear and xenophobia~\cite{TPDL_Vogel19}.

Despite growing research in this area, important gaps remain. Most available datasets are focused on English and U.S.-centric, which limits their relevance for studying European contexts~\cite{systematichp}. Moreover, few works connect hyperpartisanship with PRCTs or relate them to stance, and none provide a multilingual benchmark focused on immigration in Europe. Existing efforts examine general propaganda~\cite{dasanmartinoSemEval2020Task112020}, expand to broader categories~\cite{piskorski-etal-2023-multilingual}, or target specific issues such as climate change~\cite{Maggini}, but the combined, multilingual perspective on European migration debates is still missing.

To address these gaps, we introduce \textsc{PartisanLens}, a dataset of \num{1617} news headlines on European immigration in Spanish, Italian, and Portuguese, collected from \num{329} outlets between 2020 and 2024 via Media Cloud\footnote{\url{https://www.mediacloud.org/}}. Each headline is annotated for hyperpartisanship, Population Replacement Conspiracy Theories (PRCT), stance, and three types of rhetorical bias: loaded language, appeal to fear, and name-calling. We conduct experiments to assess the classification performance of transformer-based models and LLMs, establishing new baselines for this task. In addition, we evaluate the potential of LLMs as surrogate annotators, analysing their ability to replicate human judgments. Finally, we design a novel \textit{persona simulation} experiment, in which LLM annotators are conditioned on socio-economic and ideological profiles (derived from political compass and prism tests) to emulate different annotator backgrounds. We compare the agreement between simulated and real annotators, offering new insights into how ideological perspectives shape annotation.

Our main contributions are threefold. $(i)$ We release \textsc{PartisanLens}, the first multilingual dataset on hyperpartisan and conspiratorial immigration narratives including languages that are under-represented in political-narrative detection research. $(ii)$ We propose a joint annotation scheme that integrates hyperpartisanship, PRCTs, stance, and rhetorical bias, capturing interconnected dimensions of political discourse. $(iii)$ We conduct extensive LLM-based experiments, including the study of ideology-conditioned user-personas, showing both the promise and limitations of LLMs for subjective and context-dependent tasks. All data, code, and prompts are released publicly to support further research on multilingual misinformation detection\footnote{Dataset and code are available here: \url{https://github.com/MichJoM/PartisanLens}.}.

\section{Related Work}
\paragraph{Resources for multilingual political misinformation detection}
The growing presence of misinformation and polarised political discourse has driven extensive efforts to curate datasets for detection tasks. Prior work has focused on identifying conspiracy theories mostly on tweets covering the COVID-19 outbreaks and following episodes. \citet{Pogorelov} collected tweets relating to COVID-19 with 5G wireless network for misinformation detection. By expanding the topics in relation with COVID-19, \citet{langguth2023coco} introduced 12 categories like Satanism, Behaviour control, Intentional Pandemic, collecting conspiracy narrative tweets across different countries from 2020 to 2021. They also provided a list of actors, or perpetrators, of the pandemic narrative. With a particular focus on middle-income countries like Brazil, Indonesia and Nigeria, \citet{kim-etal-2023-covid} investigated the tweets on COVID-19 vaccine misinformation from 2020 to 2022. Lastly, \citet{miani2021loco} released the LOCO topic-matched corpus, fathered via ready-made lists of conspiracy and mainstream websites. 

In the related domain of hyperpartisan detection, \citet{kiesel-etal-2019-semeval} is a fundamental work, since it was a highly participated shared task working on English news articles covering different U.S. based topics (e.g. presidency, and gun shootings). Given the nuanced nature of hyperpartisan language, related work often includes the detection of Fake News and Satire. \citet{Golbeck} collected a dataset of satirical (203) and fake news (283) stories on American politics. Despite these advances, many existing resources remain limited in both linguistic and topical coverage. Most high-quality datasets are in English and focus on U.S.-centric themes. Indeed, few works focus on scarce resource languages like Bangla, German and Italian \cite{hossain-etal-2020-banfakenews, TPDL_Vogel19, Maggini}.

Stance detection, which involves classifying support, opposition or neutrality toward a controversial or polarising topic \cite{mohammad2017stance}, has also seen significant resource creation. \citet{mohammad-etal-2016-semeval} used tweets to track people's utterances towards Climate Change, Feminism, Donald Trump and other topics. \citet{luo-etal-2020-detecting} curated a dataset of 56k news articles on global warming seen from a U.S. perspective. Lastly, \citet{Mets} annotated Estonian articles challenging the immigration stance detection, comparing GPT-3.5 and BERT-based classifiers.

While recent studies have begun to explore extensive multilingual misinformation like \citet{Nielsen} and \citet{Leite} incorporating more than 40 languages, important gaps persist. In particular, there is still a shortage of high-quality multilingual datasets that reflect the European context like \citet{mohtaj2024newspolyml} and jointly address hyperpartisanship, stance, and rhetorical bias as interconnected elements of political discourse. 

\paragraph{LLM as annotators}
Recent advances show that political science has increasingly adopted supervised machine learning as a powerful tool for large-scale analysis of political text \cite{horych-etal-2025-promises, Stromer}. Within this context, LLM-based classification offers a promising way to reduce manual annotation effort and costs while maintaining high levels of accuracy. 

\citet{Tornberg} adopted LLMs to annotate messages from the social media platform \emph{X} based on the political belonging of the poster across 11 languages (only one from the Romance group) using the Twitter Parliamentarian Database \cite{van2020twitter}. When comparing their results with the gold-standard labels, they found that LLMs outperformed expert coders and fine-tuned baselines like BERT. However, results on public corpora may be inflated by pretraining contamination (data leakage) if models were exposed to the same sources during pretraining~\cite{sainz-etal-2023-nlp}. \citet{Heseltine} report that GPT-4 performance drops for more abstract tasks (e.g., ideology detection) and for longer inputs (full articles), while simple binary decisions yield higher human–model agreement; they also note lower accuracy and increased latency for non-English texts.

%Finally, the implementation choices made by researchers while using LLMs as an annotator influence the results, introducing potential systematic biases and random errors \cite{baumann2025largelanguagemodelhacking}. 
Therefore, further investigation is needed to understand how incorporating additional inputs like sociodemographic data might enhance the performance of LLMs as annotators in subjective NLP tasks \cite{beck-etal-2024-sensitivity}.

Finally, the choices made when using LLMs as annotators can introduce biases and errors \cite{baumann2025largelanguagemodelhacking}; recent works show that LLMs are not fully reliable at the instance level in sensitive tasks, yet capture overall trends \cite{piot2025llmsevaluateannotaterevisiting}. Therefore, further investigation is needed to see how additional inputs, such as sociodemographic data, might improve LLM annotation in subjective NLP tasks \cite{beck-etal-2024-sensitivity}.

%Overall, \textcolor{red} current research reflects a growing consensus on the effectiveness of LLMs as data annotators} \cite{pavlovic-poesio-2024-effectiveness}. %However, important questions remain about how LLMs perform on culturally and linguistically complex political annotation tasks, particularly in multilingual European contexts where cultural nuances significantly impact interpretation.

\paragraph{LLM persona}

Variation among human annotators has long been recognised as a source of noise against a presumed \emph{ground truth} \cite{plank-2022-problem, schafer-etal-2025-demographics}. This issue is especially evident in subjective tasks such as toxic or racist language detection \cite{sap-etal-2022-annotators} and natural language inference \cite{biester-etal-2022-analyzing}, where sociodemographic differences strongly influence interpretation. Recent research examines how real or synthetic demographic profiles (e.g., PersonaHub \cite{ge2024scaling}) correlate with annotation behaviour. In this direction, \citet{frohling-etal-2025-personas} showed that injecting persona descriptions into LLM prompts enhances diversity and control in annotation, while \citet{mukherjee-etal-2024-cultural} explored cultural variation through persona-based elicitation. Complementing these studies, \citet{piot2025personalisationprejudiceaddressinggeographic} demonstrated that geographic bias affects how state-of-the-art LLMs classify hate speech, revealing the influence of user-personas on model outputs. Understanding such variations is essential when assessing whether LLMs can replicate or complement human annotation in politically sensitive domains, where models remain highly sensitive to design choices and vulnerable to manipulation or \emph{LLM hacking} \cite{baumann2025largelanguagemodelhacking}.

\section{Dataset Construction}

In this section, we present the data collection scheme, the annotation protocol and the resulting dataset statistics.

\subsection{Data collection and preparation}\label{sec:DatasetCollection}

We focused on collecting immigration-related headlines in three underrepresented Romance languages: Spanish, Portuguese and Italian. These represent major linguistic areas along the Mediterranean migration routes into Europe \cite{iemed2018Migration}, and yet remain largely absent from resources for misinformation detection. Our data collection followed a hybrid strategy that combined keyword-based retrieval with LLM-assisted filtering. Using the Media Cloud platform\footnote{\url{https://www.mediacloud.org/}}, we extracted headlines from national and regional outlets in Spain, Italy, and Portugal. For each language, we used targeted keyword queries (e.g., equivalents of \emph{migration} or \emph{immigration policy}, \emph{population replacement}) to capture a broad range of immigration narratives\footnote{Full keyword lists are available in our repository.}. After removing duplicate URLs and headlines, this process resulted in approximately \num{400,000} unique entries.

Given this scale, we employed \texttt{Llama 3.1 8B} to perform an initial classification along two dimensions: hyperpartisan and PRCT. This automatic step ensured coverage of low-frequency but socially important categories. In particular, PRCT-related content represented less than 1\% of the initial corpus. Guided by this filtering, we selected a balanced and diverse sample of \num{1617} headlines published between 2020 and 2024 from \num{329} distinct sources.

\subsection{Annotation protocol}\label{sec:Annotation_Protocol}

The annotation protocol was designed to capture both ideological content and rhetorical framing in immigration-related headlines. Each item was annotated along \emph{six} complementary dimensions that reflect the stance, narrative strategy, and emotional tone of the text\footnote{Full annotation guidelines are available in our repository.}. At the content level, we considered: $(i)$ \textbf{Hyperpartisan}, indicating whether the headline expressed a one-sided or ideologically extreme view; $(ii)$ \textbf{PRCT}, marking conspiratorial framings such as references to population replacement or coordinated demographic change; and $(iii)$ \textbf{Stance} (categorical), capturing the overall attitude toward immigration as \emph{pro}, \emph{neutral}, or \emph{against}. 

In addition, three rhetorical strategies were annotated at the span level: $(iv)$ \textbf{Loaded language}, where emotionally charged wording was used to provoke reaction; $(v)$ \textbf{Appeal to fear}, framing immigration as a threat or danger; and $(vi)$ \textbf{Name calling}, involving derogatory or stereotyping expressions. These rhetorical cues are known to polarise debate by appealing to emotion rather than reasoned argumentation~\cite{Maggini}.

Annotation was conducted through a custom Web interface that masked outlet names to avoid source bias. Nine native speakers, 3 males and 6 females (three per language, all with at least a master's degree) carried out the task over four iterative rounds, beginning with pilot annotations and guideline refinement, followed by progressively larger batches. In total, the process required approximately \num{66} hours. The final dataset includes both individual annotations and majority labels, enabling not only supervised training but also future research on annotator disagreement and uncertainty.

To assess the annotation reliability, we computed Fleiss' $\kappa$ across the three main annotation tasks (Hyperpartisan, PRCT, Stance). We did not compute agreement for the rhetorical strategies, as these were annotated at the span level. Their purpose was primarily to guide annotators by highlighting cues associated with hyperpartisan phenomena. Results are presented in Table~\ref{tab:dataset_agreement} and indicate substantial to almost perfect agreement across all languages~\cite{Landis1977}. Hyperpartisan labels were most consistent in Portuguese and Spanish, while PRCT achieved slightly higher agreement in Italian. Stance annotations were somewhat lower than the other tasks but still showed substantial reliability.

\begin{table}[t]
    \centering
    \footnotesize
    \begin{tabular}{lrrr}
    \toprule
    & \textbf{SPA} & \textbf{ITA} & \textbf{PT} \\
    \midrule
    \textbf{Hyperpartisan} & 0.876 & 0.706 & 0.770 \\
    \textbf{PRCT} & 0.880 & 0.774 & 0.721 \\
    \textbf{Stance} & 0.837 & 0.744 & 0.737 \\
    \bottomrule
    \end{tabular}
    \caption{Inter-Annotator Agreement (Fleiss' $\kappa$).}    \label{tab:dataset_agreement}
\end{table}

\subsection{Dataset statistics}

\textsc{PartisanLens} includes 527 headlines in Portuguese (PT), 565 in Italian (ITA), and 525 in Spanish (SPA). Table~\ref{tab:dataset_comprehensive} summarises the majority voting distribution across all labels. Hyperpartisan headlines are prevalent in every language (range of 59-70\%), while PRCT cases remain relatively rare (under 10\% in PT and SPA, but higher in ITA at 28\%). Stance distribution varies considerably: anti-immigration headlines dominate in Italian (59\%) and Portuguese (40\%), whereas Spanish headlines are mostly neutral (80\%). Pro-immigration stances are scarce across the three languages.

\begin{table}[t]
  \centering
  \small
  \resizebox{\columnwidth}{!}{%
  \begin{tabular}{@{}l l r r r}
    \toprule
    \textbf{Category} & \textbf{Label} & \textbf{SPA} & \textbf{ITA} & \textbf{PT} \\
    \midrule
    \multirow{2}{*}{Hyperpartisan} & True & 336 & 337 & 367 \\
                                 & False & 189 & 228 & 160 \\
    \midrule
    \multirow{2}{*}{PRCT}    & True  & 46 & 160 &  9 \\
                                 & False & 479 & 405 & 518 \\
    \midrule
    \multirow{3}{*}{Stance}  & Pro      & 4 &  78 &   130 \\
                              & Neutral   & 419 & 156 & 185 \\
                              & Against   & 102 & 331 & 212 \\
    \midrule
    \multirow{2}{*}{Loaded lang.} & Present  & 346 & 234 & 213 \\
                              & None      & 179 & 331 & 314 \\
    \midrule
    \multirow{2}{*}{Appeal to fear} & Present &  107 & 168 & 35 \\
                                   & None    & 418 & 397 & 492 \\
    \midrule
    \multirow{2}{*}{Name calling}  & Present &  35 &  44 &  33 \\
                                   & None    & 490 & 521 & 494 \\
    \bottomrule
  \end{tabular}%
  }
  \caption{Dataset label distribution.}
  \label{tab:dataset_comprehensive}
\end{table}

\begin{figure}[h]
  \centering
  \includegraphics[width=0.9\columnwidth]{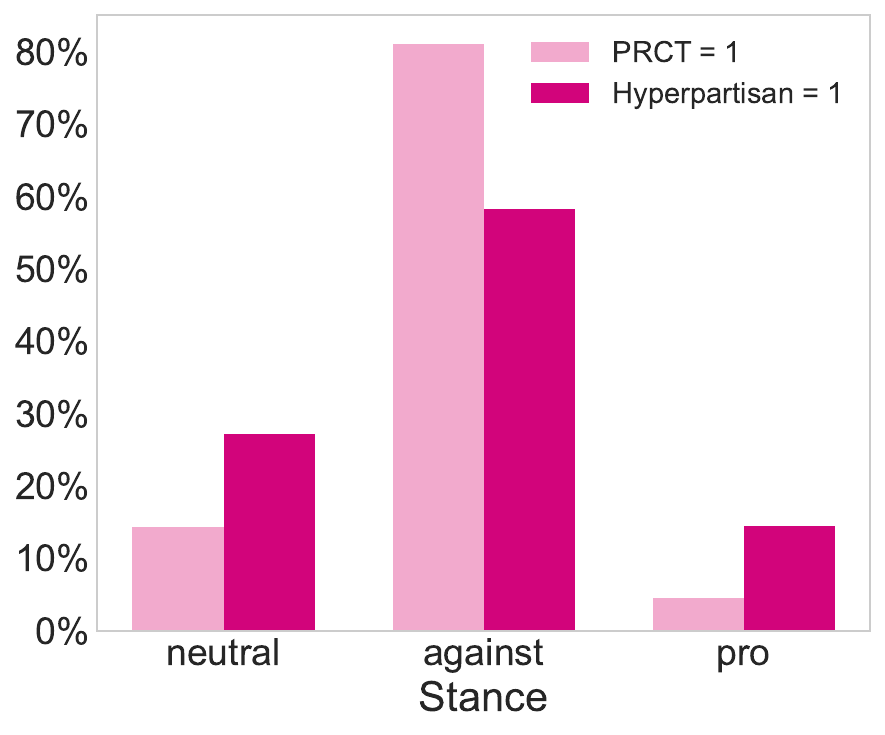}
    \caption{Hyperpartisan and PRCT stance proportions.}
    \label{fig:prct_stance_hyperpar}
\end{figure}

% me parece muy raro que haya alguna noticia de PRCT y que esté a favor de la immigración, no tiene sentido, pero bueno
\textbf{Stance-conditioned patterns.}  
Figure~\ref{fig:prct_stance_hyperpar} shows how PRCT and hyperpartisan content correlate with stance. The strongest signal comes from anti-immigration headlines: 81\% contain PRCT claims and 56\% are hyperpartisan. Neutral headlines are minor (14\% PRCT, 27\% hyperpartisan), while pro-immigration headlines are the least polemical (3\% PRCT, 12\% hyperpartisan). These results confirm that conspiratorial narratives cluster much more around negative stances than hyperpartisan style, which is more broadly distributed.

\begin{figure}[h]
  \centering
  \includegraphics[width=0.9\columnwidth]{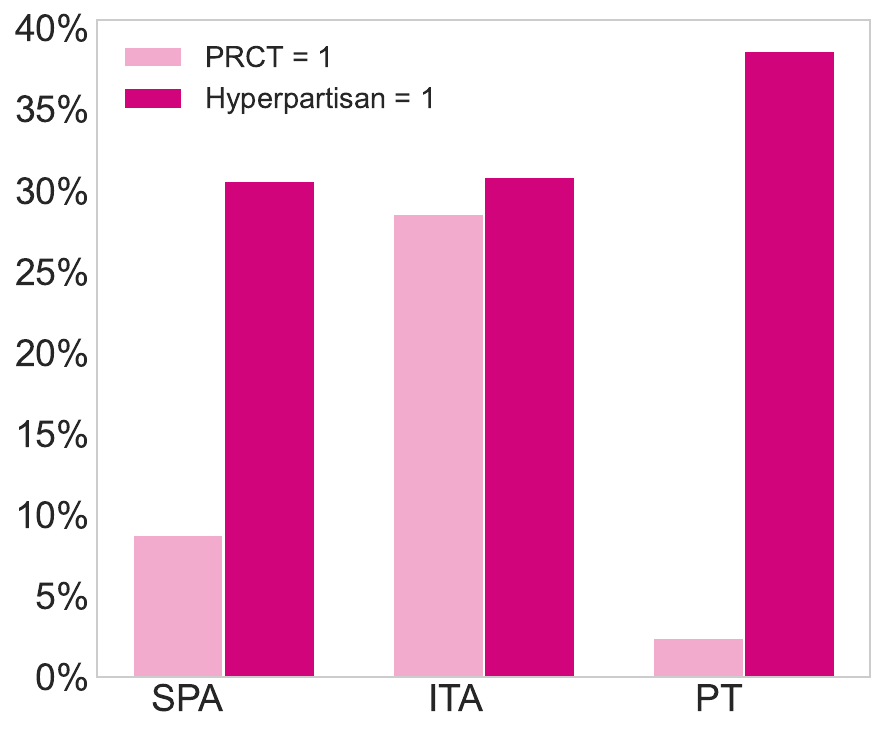}
    \caption{Hyperpartisan and PRCT content by language.}
    \label{fig:hyperpartisan_media}
\end{figure}

\paragraph{\textbf{Language‐level variation}}
Figure \ref{fig:hyperpartisan_media} highlights how the two labels distribute across the languages' headlines. Italian outlets display the strongest conspiratorial signal: 27\% of their immigration headlines contain PRCT language, roughly three times the Spanish rate (8\%) and nine times the Portuguese rate (3\%). Hyperpartisan style shows a different trend: it peaks in Portuguese (70\%) and remains consistently high in Spanish and Italian ($\approx$60\%). Taken together, these findings suggest that while hyperpartisan framing is common across languages, conspiratorial narratives concentrate more on Italian media.

\section{Benchmark Experiments: Setup and Results}
\label{sec:Comparison}

To assess the utility of \textsc{PartisanLens} as a benchmarking resource, we compare state-of-the-art multilingual models under increasing supervision levels, from zero-shot prompting to supervised fine-tuning. Our goals are: (i) to establish baselines for the three target tasks (Hyperpartisan, PRCT, Stance) across Spanish,  Italian and Portuguese (SPA, ITA, PT), and (ii) to quantify how supervision type and rationale-augmented learning affect accuracy and robustness.

\subsection{Models}

We evaluate multilingual transformers covering encoder and generative architectures. As a strong non-generative baseline, we include \texttt{mBERT}~\cite{devlin2019bertpretrainingdeepbidirectional} (110M parameters), widely used for cross-lingual classification. To compare against more recent instruction-tuned systems, we explore three generative LLMs of increasing scale: \texttt{Llama 3.1–8B}, \texttt{Mistral–Nemo–Base–2407} (12B) \cite{mistral_nemo} and \texttt{Llama 3.3–70B} \cite{dubey2024llama3herdmodels}.

\subsection{Classification strategies}

We design four complementary strategies to study supervision effects over the three target tasks (hyperpartisan, PRCT and stance):

\paragraph{\textbf{Zero-shot prompting.}}
Models receive a concise task description defining the three task categories. We set $\texttt{temperature}=0$ for deterministic decoding and instruct models to output only a valid label (see Appendix, Table~\ref{tab:zero-shot-prompt}).

\paragraph{\textbf{Few-shot prompting.}}
To provide minimal task-specific context, we include ten balanced demonstrations ($\approx$2 per label) selected via a Determinantal Point Process (DPP)~\cite{Kulesza_2012} to maximise diversity and reduce redundancy. We interleave one positive instance and one negative in our few-shot. Prompts match the zero-shot setup except for the in-context examples. Further DPP details appear in Appendix~\ref{sec:APP_DPP}.

\paragraph{\textbf{Supervised fine-tuning (labels).}}
LLMs are fine-tuned for two epochs with learning rate $2e^{-5}$, linear decay, warm-up ratio 0.03, batch size 12, and \texttt{adamw\_8bit}; inputs are truncated at 4096 tokens. These hyperparameters follow common practice in instruction-tuning and were selected after small validation sweeps to balance stability and avoid overfitting. For \texttt{mBERT}, we train for 5 epochs with learning rate $3e^{-4}$, batch size 16, and dropout 0.23.

\begin{figure}[t]
  \centering
  \includegraphics[width=\columnwidth]{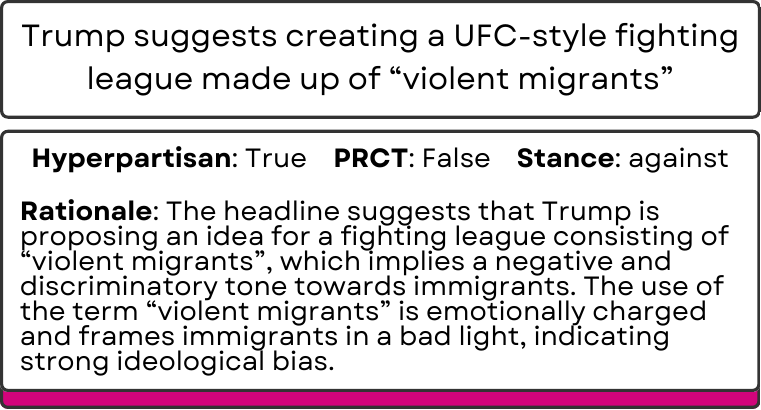}
  \caption{Example of a rationale-enhanced annotation.} % including a news headline, labels and natural-language rationales generated from annotated rhetorical strategies
  \label{fig:rationale_example}
\end{figure}

\paragraph{\textbf{Supervised fine-tuning (labels $+$ rationales).}}
In this variant, models are fine-tuned not only on classification labels but also on rationales, which are short, natural-language explanations derived from the rhetorical bias spans annotated during corpus creation. These rationales express how a headline carries persuasion, such as: \textit{``The headline appeals to fear by suggesting a national threat''}. We generate fluent, context-aware rationales from annotated templates using \texttt{Llama-3.3-70B-Instruct}, and concatenate them to the input text during fine-tuning. This encourages interpretable learning and alignment between evidence and prediction \cite{Zhang_Wu_Xu_Cao_Du_Psounis_2024,Piot2025}. Example of a rationale-enhanced instance is shown in Figure~\ref{fig:rationale_example}.

\subsection{Evaluation metrics}

We report macro-averaged precision, recall, and F1 across the three tasks and all our proposed models. These metrics are robust to class imbalance and ensure that minority classes contribute equally to the overall score. This follows standard practice in hyperpartisan detection, misinformation classification, and abusive-language research, where label distributions are typically skewed \cite{liu2024mmfakebench, essahli2025fakezerorealtimeprivacypreservingmisinformation, assenmacher2025explicitbilingualdatasetdehumanization}.

\subsection{Results}

Table~\ref{tab:experiments} reports macro-averaged precision (P), recall (R), and F1 across the three tasks and all our proposed models.

\begin{table*}[ht]
\centering
\small
\setlength{\tabcolsep}{5pt}
    \begin{tabular}{ll rrr rrr rrr}
    \toprule
    \multirow{2}{*}{\textbf{Model}} & \multirow{2}{*}{\textbf{Setting}} &
    \multicolumn{3}{c}{\textbf{Hyperpartisan}} &
    \multicolumn{3}{c}{\textbf{PRCT}} &
    \multicolumn{3}{c}{\textbf{Stance}} \\
    \cmidrule(lr){3-5}\cmidrule(lr){6-8}\cmidrule(lr){9-11}
    & & P & R & F1 & P & R & F1 & P & R & F1 \\
    \midrule
    \texttt{mBERT} & fine-tuned & 0.6896 & 0.6312 & 0.6299 & 0.8589 & 0.8148 & \textbf{0.8345} & 0.6142 & 0.6047 & 0.5974 \\
    \midrule
    \multirow{4}{*}{\texttt{Llama 8B}} 
     & zero-shot     &  0.6483 & 0.6483 & 0.6483 & 0.7419 & 0.6056 & 0.6306 & 0.5795 & 0.4569 & 0.4264 \\
     & few-shot      & 0.7117 & 0.7165 & \textbf{0.7129} & 0.7274 & 0.8041 & 0.7540 & 0.6227 & 0.5768 & 0.5761 \\
     & fine-tuned    & 0.6863 & 0.6662 & 0.6689 & 0.7751 & 0.6980 & 0.7263 & 0.6121 & 0.4846 & 0.4639 \\  
     & rationales   & 0.6973 & 0.6680 & 0.6706 & 0.7513 & 0.6762 & 0.7028 & 0.6491 & 0.4674 & 0.4320 \\
     \midrule
    \multirow{4}{*}{\texttt{Llama 70B}} 
     & zero-shot     & 0.6806 & 0.6885 & 0.6654 & 0.7814 & 0.8537 & 0.8090 & 0.6273 & 0.6042 & 0.5945  \\
     & few-shot      & 0.7054 & 0.6903 & 0.6938 & 0.8281 & 0.8175 & 0.8226 & 0.6457 & \textbf{0.6112} & 0.5964 \\
     & fine-tuned    & 0.7047 & 0.7085 & 0.6911 & 0.7607 & \textbf{0.8654} & 0.7943 & 0.6413 & 0.6086 & 0.6062 \\
     & rationales   & \textbf{0.7179} & \textbf{0.7246} & 0.7118 & 0.7295 & 0.8471 & 0.7598 & 0.6709 & 0.6051 & 0.6068 \\ 
     \midrule
    \multirow{4}{*}{\texttt{Nemo}} 
     & zero-shot     & 0.6818 & 0.6839 & 0.6827 & \textbf{0.9275} & 0.7111 & 0.7668 & 0.5580 & 0.4750 & 0.4433 \\
     & few-shot      & 0.7039 & 0.7056 & 0.6851 & \textbf{0.9325} & 0.7237 & 0.7814 & 0.6717 & 0.5995 & \textbf{0.6138} \\
     & fine-tuned    & 0.7061 & 0.7109 & 0.7073 & 0.8875 & 0.7004 & 0.7512 & 0.5977 & 0.4849 & 0.4625 \\
     & rationales   & 0.6843 & 0.6474 & 0.6475 & 0.8117 & 0.6735 & 0.7126 & \textbf{0.6877} & 0.4642 & 0.4232 \\ 
     \bottomrule
    \end{tabular}
    \caption{Macro-averaged precision (P), recall (R), and F1 across tasks and all our considered models. Best metrics per task are marked in \textbf{bold}.}
    \label{tab:experiments}
\end{table*}

\textbf{Hyperpartisan.} Performance is solid but not the easiest task (F1$\approx$0.62–0.71). Top scores come from \texttt{Llama~8B} (few-shot), \texttt{Llama~70B} (rationales), and \texttt{Nemo} (fine-tuned), all around $0.71$ with balanced precision and recall. Scale helps, but real demonstrations let small models match larger ones. Rationale inputs benefit \texttt{Llama~70B}, suggesting that style-sensitive evidence can be exploited at large capacity, but the other architectures do not share this pattern. Overall, hyperpartisan detection benefits most from light supervision, with few-shot prompts often matching, or even replacing, the need for large-scale fine-tuning.

\textbf{PRCT.} Detecting PRCT is \emph{easier} than hyperpartisan related content. The best F1 is $0.82$ (\texttt{Llama~70B} few-shot), and even zero-shot \texttt{Llama~70B} is strong (F1=$0.81$). \texttt{Nemo} achieves the highest precision (0.93) at the cost of recall ($\approx$0.71–0.72), pointing to a conservative detector that captures clear conspiracy markers but misses subtler instantiations. Notably, \texttt{mBERT} fine-tuned also reaches F1=$0.82$. Rationale augmentation does not help here, aligning with the fact that our rationales encode rhetorical style rather than conspiracy semantics. The limiting factor is recall, not interpretability alignment.

\textbf{Stance.} Stance is the hardest task (best F1=$0.61$), with \texttt{Llama~70B} (fine-tuned / rationales) and \texttt{Nemo} (few-shot) tying at the top. Gains mostly come from modest recall ($0.61$), while smaller models stay below (F1=$0.58$). The pronounced difficulty likely reflects both class distributions (e.g., very rare pro stance in Spanish) and the need for contextual inferences beyond surface style, since rationale inputs can even hurt (with Nemo obtaining a F1=$0.42$). The overall difficulty suggests that stance requires finer-grained semantic understanding than hyperpartisan or PRCT, and remains an open challenge even for very large models.

Three main insights emerge from these results. ($i$) Supervision matters: few-shot and fine-tuning typically add 4–7 F1 points over zero-shot. ($ii$) Scale helps but is not enough: \texttt{Llama~70B} is most consistent, yet \texttt{Llama~8B} few-shot matches top hyperpartisan scores. ($iii$) Task complexity varies: \emph{PRCT $>$ hyperpartisan $>$ stance}. PRCT benefits from distinctive lexical/conceptual signatures, hyperpartisan is stylistic and diffuse, and stance seems to demand finer pragmatic cues. A detailed per-language analysis, including F1\textsubscript{MACRO}, P/R, and error profiles, is provided in Appendix~\ref{app:lang_analysis}.

\section{LLMs as Potential Data Annotators}
\label{sec:llms_annotators}

To evaluate whether LLMs can serve as reliable annotators, we compare their agreement against the reference level of human agreement across languages. The core question is: \emph{does an LLM approach human agreement levels on our three targets (Hyperpartisan, PRCT, Stance) in the three languages?}

\subsection{Setup}

% For this reason, we measure mean pairwise Cohen’s~$\kappa$ for each human–human (H-H) and for LLM–human (L-H) pairs, using a single strong model (\texttt{Llama~3.3--70B} fine-tuned with rationales). We selected \texttt{Llama 3.3 70B} fine-tuned \textcolor{blue}{using the labels $+$ rationales setup (i.e. trained on both the headline and a natural-language rationale describing its persuasive strategy)} because we believe it could leverage its internal political knowledge better than smaller models and because it resulted in being robust across most tasks. The language H-H and L-H agreements are summarized in Table~\ref{tab:kappa_means}\footnote{Each H-H value averages across all human pairs, and the L-H and G-H values average the LLM vs.\ each human.}.

To answer this research question, we measure mean pairwise Cohen’s~$\kappa$ for human–human (H--H) and LLM–human pairs using two models: (\textit{i}) \texttt{Llama~3.3--70B} fine-tuned with labels\,+\, rationales (L--H), and (\textit{ii}) \texttt{gpt-5-mini} in inference mode (G--H). We chose the 70B model with rationales due to its robustness across tasks, and include \texttt{gpt-5-mini} to study whether a recent inference model narrows the human gap. The agreement results for all the annotation pairs (human pairs and LLM-human pairs) are summarised in Table~\ref{tab:kappa_means_combined}\footnote{Each H--H value averages across all human pairs. L--H and G--H average the corresponding LLM vs.\ each human.}.

\subsection{Results}

%\textcolor{red}{TODO: update with values from the rerun for camera ready}
\begin{table}[t]
  \centering
  \scriptsize
  \setlength{\tabcolsep}{2pt}
  \renewcommand{\arraystretch}{0.95}
  \begin{tabular}{@{}l
                  rrr
                  rrr
                  rrr@{}}
    \toprule
    \multirow{2}{*}{\textbf{Label}} &
    \multicolumn{3}{c}{\textbf{Spanish}} &
    \multicolumn{3}{c}{\textbf{Italian}} &
    \multicolumn{3}{c}{\textbf{Portuguese}} \\
    \cmidrule(lr){2-4}\cmidrule(lr){5-7}\cmidrule(lr){8-10}
    & \textbf{H--H} & \textbf{L--H} & \textbf{G--H}
    & \textbf{H--H} & \textbf{L--H} & \textbf{G--H}
    & \textbf{H--H} & \textbf{L--H} & \textbf{G--H} \\
    \midrule
    Hyperpartisan & 0.864 & 0.424 & 0.594 & 0.698 & 0.524 & 0.452 & 0.666 & 0.228 & 0.219 \\
    PRCT          & 0.893 & 0.483 & 0.549 & 0.743 & 0.632 & 0.541 & 1.000 & 0.124 & 0.662 \\
    Stance        & 0.858 & 0.364 & 0.633 & 0.598 & 0.377 & 0.567 & 0.682 & 0.392 & 0.480 \\
    \bottomrule
  \end{tabular}
  \caption{Mean Cohen’s $\kappa$ by language and label. The agreement pairs are: \textbf{H--H} (human vs. human), \textbf{L--H} (\texttt{Llama~3.3--70B} vs.\ human), \textbf{G--H} (\texttt{gpt-5-mini} vs.\ human).}
  \label{tab:kappa_means_combined}
\end{table}

\begin{comment}
\begin{table}[t]
  \centering
  \small
  % slightly tighter column padding so the wider table fits
  \setlength{\tabcolsep}{4pt}
  \begin{tabular}{@{}l r r r r r r@{}}
    \toprule
    \multirow{2}{*}{\textbf{Label}} &
    \multicolumn{2}{c}{\textbf{Spanish}} &
    \multicolumn{2}{c}{\textbf{Italian}} &
    \multicolumn{2}{c}{\textbf{Portuguese}} \\
    \cmidrule(lr){2-3}\cmidrule(lr){4-5}\cmidrule(lr){6-7}
      & P-P & L-P & P-P & L-P & P-P & L-P \\
    \midrule
    Hyperpartisan & 0.928 & 0.612 & 0.972 & 0.734 & 0.937 & 0.698 \\
    PRCT          & 0.877 & 0.745 & 0.967 & 0.775 & 0.860 & 0.563\\
    Stance        & 0.933 & 0.742 & 0.943 & 0.769 & 0.949 & 0.661 \\
    \bottomrule
  \end{tabular}
  \caption{Mean Cohen’s $\kappa$ agreement. P-P stands for LLM-personas pairs and L-P for LLM versus LLM-personas.}
  \label{tab:kappa_means_llm_persona}  
\end{table}
\end{comment}

\paragraph{\textbf{Human agreements (H-H)}}
Among the three tasks, human agreement is overall higher for Spanish headlines ($\kappa \approx 0.86$–$0.88$), followed by Italian ($\kappa \approx 0.59$–$0.74$), with more variability in Portuguese ($\kappa \approx 0.66$–$1.00$). We attribute the perfect agreement for PCRT in Portuguese due to extreme class sparsity (only $n{=}9$ positives). The Spanish annotators higher agreement, which may be related to the country's generally higher level of immigration integration and a more informed understanding of immigration issues, as reported by the \citet{oecdIndicatorsImmigrant}'s Indicators of Immigrant Integration 2023. This pattern suggests that both linguistic factors and local media influence how easy it is to judge a headline.

\paragraph{\textbf{LLM vs.\ human agreements (L--H / G--H)}}

Models remain below H--H across languages, but \texttt{gpt-5-mini} (G--H) is closer to human agreement than \texttt{Llama~3.3--70B} (L--H). In Spanish, \texttt{gpt-5-mini} improves over \texttt{Llama~3.3--70B} on all tasks, especially on the Stance category ($0.633$ vs. $0.364$). In Italian, however, \texttt{Llama~3.3--70B} obtains better results in Hyperpartisan and PRCT than \texttt{gpt-5-mini}, but \texttt{gpt-5-mini} still obtains way better performance on Stance, being close to human agreement ($0.567$ vs. $0.598$). In Portuguese, \texttt{gpt-5-mini} yields the large PRCT difference in comparison to  \texttt{Llama~3.3--70B} ($0.662$ vs.\ L--H $0.124$), but, in this case, for the three labels, the LLMs are still way below compared to human agreement.

In sum, although we evaluated two large and competitive LLMs, its limited agreement with experts shows that fully unsupervised annotation remains out of reach for these nuanced categories. Human supervision is therefore still indispensable, making resources in this domain crucial for advancing robust detection methods.

\section{User-Persona LLMs}

Persona-Based prompting intends to condition a models’ output to reflect the characteristics of specific personas, enabling researchers to simulate opinions, values, and attitudes more effectively~\cite{beck-etal-2024-sensitivity}. In this experiment, we examine whether LLMs can emulate patterns of human annotation behaviour when conditioned on realistic user profiles. Our motivation is to assess whether user-persona LLMs reflect the same socio-demographic and ideological variability observed among human annotators. In other words, if human disagreement is partially driven by background traits, an LLM simulating those traits should align more closely with annotators sharing the same profile. We therefore ground LLM personas using real socio-demographic and political data from our annotators, and compare agreement levels between persona-conditioned LLMs and their human counterparts across Spanish, Italian, and Portuguese.

\subsection{Persona prompt}
We build structured persona prompts representing each annotator’s political and demographic background. To do so, each human annotator completed two anonymous instruments: the Political Compass test,\footnote{\url{https://www.politicalcompass.org/}} which places respondents on Left–Right and Authoritarian–Libertarian axes, and the PRISM quiz,\footnote{\url{https://prismquiz.github.io/}} which captures social and economic orientations. We standardise test scores within each dimension and combine them with self-reported sociodemographics (e.g., age range, gender) to construct concise, non-identifying textual profiles. Personas are generated from predefined templates that map score ranges to qualitative descriptors (e.g., “economically left-leaning, socially moderate, libertarian-leaning”). Full sociodemographic summaries and persona templates are included in the Tables~\ref{tab:sociodemodata} and~\ref{tab:zero-shot-prompt-persona} (Appendix).

\subsection{Setup}

We implement persona prompting with two models: \texttt{Llama~3.1--70B} and \texttt{gpt5-mini}. For both, we extend the zero-shot classification prompt from Section~\ref{sec:Comparison} with the same user-persona context, with all other prompt components remaining identical. The persona context states the annotator’s traits (gender, age range, and economic/social/authority–liberty positions derived from Compass/PRISM) before the task instructions. For evaluation, we compare each persona-LLM to: $(i)$ its matched human counterpart (the annotator whose profile it simulates) and $(ii)$ non-matched annotators in the same language. We also contrast these results with the human–human agreement baseline (Table~\ref{tab:kappa_means_combined}).

\subsection{Results}

\begin{table}[t]
    \centering
    \footnotesize
    \setlength{\tabcolsep}{3.5pt}
    \resizebox{\columnwidth}{!}{%
    \begin{tabular}{l cc cc cc}
    \toprule
    \multirow{2}{*}{\textbf{Persona}} &
    \multicolumn{2}{c}{\textbf{Hyperpartisan}} &
    \multicolumn{2}{c}{\textbf{PRCT}} &
    \multicolumn{2}{c}{\textbf{Stance}} \\
    \cmidrule(lr){2-3}\cmidrule(lr){4-5}\cmidrule(lr){6-7}
    & L70B & gpt5-m & L70B & gpt5-m & gpt5-m & gpt5-m \\
    \midrule
    SPA\_1 & 0.5331 & 0.6334 & 0.5531 & 0.6252 & 0.3441 & 0.7672 \\
    SPA\_2 & 0.4151 & 0.6351 & 0.5889 & 0.7540 & 0.2162 & 0.6192 \\
    SPA\_3 & 0.4276 & 0.5255 & 0.6550 & 0.7540 & 0.1969 & 0.6020 \\
    \midrule
    ITA\_1 & 0.4908 & 0.5068 & 0.7409 & 0.7274 & 0.3758 & 0.6316 \\
    ITA\_2 & 0.5024 & 0.5278 & 0.6217 & 0.7060 & 0.3049 & 0.5374 \\
    ITA\_3 & 0.4658 & 0.4567 & 0.7392 & 0.6320 & 0.3629 & 0.5882 \\
    \midrule
    PT\_1  & 0.2822 & 0.2143 & 0.3151 & 0.6622 & 0.4367 & 0.4284 \\
    PT\_2  & 0.2435 & 0.1397 & 0.3151 & 0.6622 & 0.4313 & 0.3989 \\
    PT\_3  & 0.2136 & 0.1806 & 0.2268 & 0.6622 & 0.4860 & 0.5599 \\
    \bottomrule
    \end{tabular}
    }    
    \caption{Cohen's $\kappa$ for each user-persona vs.\ its corresponding human annotator, comparing \texttt{Llama~3.1-70B} (L70B) and \texttt{gpt5-mini} (gpt5-m) inference.}
    %anxo stance: 0.4033
    %anxo stance: 0.2991
    %anxo stance: 0.2882
    %anxo stance: 0.3868
    %anxo stance: 0.3325
    %anxo stance: 0.3862
    %anxo stance: 0.4398    
    %anxo stance: 0.3831
    %anxo stance: 0.5176
    \label{tab:LLMPersonas}
\end{table}

\paragraph{Agreement user-persona vs.\ corresponding human}

Table~\ref{tab:LLMPersonas} reports Cohen’s~$\kappa$ between each persona-conditioned LLM and the human it simulates, using \texttt{Llama--70B} and \texttt{gpt5-mini}. If we compare the overall results to the generic L–H baselines in Table~\ref{tab:kappa_means_combined}, LLM persona conditioning helps in several settings, and \texttt{gpt5-mini} often amplifies these gains, especially for PRCT and Stance. Regarding Spanish annotators, \texttt{Llama--70B} personas improves PRCT over the generic baseline ($0.599$ vs.\ $0.483$) and slightly Hyperpartisan ($0.459$ vs.\ $0.424$), but lower Stance ($0.252$ vs.\ $0.364$). This suggests that persona conditioning may help capture content focus and priorities rather than polarity judgments. With \texttt{gpt5-mini}, agreement rises across all three: Hyperpartisan ($0.598$ vs. $0.635$), PRCT ($0.711$ vs. $0.754$), and Stance ($0.663$ vs. $0.767$).

For Italian annotators, \texttt{Llama--70B} personas shows the same trend, with clear gains on PRCT ($0.701$ vs. $0.632$ ), are near parity for Stance ($0.348$ vs.\ $0.377$), and slightly below for Hyperpartisan ($0.486$ vs. $0.524$). This pattern reinforces the idea that persona conditioning is most effective for tasks involving content relevance (PRCT), while stylistic or bias judgments are less influenced by the persona information. \texttt{gpt5-mini} keeps PRCT competitive ($0.689$ vs $0.727$) and even delivers gains in Stance ($0.586$), clearly above both the baseline ($0.377$). Hyperpartisan remains a bit below baseline. Overall, Italian sees the strongest persona-driven PRCT alignment and a notable stance improvement with \texttt{gpt5-mini}.

Finally, if we analyse the Portuguese results, despite extreme PRCT sparsity, \texttt{Llama--70B} personas improve the generic baseline on all tasks. \texttt{gpt5-mini} produces a large PRCT increase ($0.662$), a modest Stance performance ($0.462$), and a decrease for Hyperpartisan ($0.178$). This pattern suggests persona cues align well with distinctive conspiratorial markers even under sparsity, while stylistic judgments remain fragile in PT.

Looking at the overall results, persona conditioning increases LLM–human PRCT agreement in all languages (largest gains in ITA and PT). Using \texttt{gpt5-mini}, it clearly improves Stance in SPA/ITA. Hyperpartisan performance remains mixed (SPA gains, ITA near parity, and PT drops). Overall levels are still way down compared to the H–H reference (Table~\ref{tab:kappa_means_combined}), but grounding in real profiles recovers meaningful human-like variability for PRCT and, with stronger inference, also for stance.

\subsection{Error analysis}
\label{sec:error_analysis_main}

\begin{table*}[t]
\small
\centering
    \begin{tabular}{p{0.47\textwidth} p{0.47\textwidth} }
    \toprule
    \textbf{Original Language} & \textbf{English Translation} \\ \midrule
    Migranti, il M5s: ``Lollobrigida parla di sostituzione etnica? E' ignorante, mica razzista'' & Migrants: M5s: ``is Lollobrigida talking about etnic replacement? He is ignorant, not racist''\\ \midrule
    El PP prepara toda su artillería contra la ``independencia fiscal'' de Cataluña & PP is preparing the artillery against the ``financial independence'' of Catalonia \\ \midrule
    Immigrazione, Trump: ``Avvelena il sangue del nostro Paese'' & ``Immigration, Trump: It poisons our Country's blood''\\ 
    \bottomrule
    \end{tabular}
\caption{Selected headlines with human and user-personas disagreements. }\label{tab:examples}
\end{table*}
\normalsize

We explore LLM's deficiencies and attribute them to several factors. One important issue is the limited understanding of sarcastic and metaphorical sentences. In Table~\ref{tab:examples}, we can see some examples of headlines with disagreements among annotators and user-personas. A second limitation concerns the model's insufficient understanding of semantics and context. For example, the Portuguese term \emph{Imigrantes} can denote both ``immigrants'' and the name of a Brazilian highway. The model struggled to disambiguate the meaning based on context. Similarly, the word ``invading'' carries different connotations across Portuguese, Italian, and Spanish, and the LLM's rigid interpretation suggests a lack of sensitivity to cultural context. Another illustrative shortcoming is the model's failure to interpret irony, sarcasm, or implicit critique. Such use of rhetorical devices is a subtle but powerful tool in political discourse, where it frequently conveys unstated or indirect opinions.

\section{Conclusions and Future Work}
\label{sec:Conclusion}

This paper introduced \textsc{PartisanLens}, a multilingual dataset of \num{1617} news headlines for analysing hyperpartisan content, population replacement conspiracy theories, and stance towards immigration news in Spanish, Italian, and Portuguese. A robust protocol with nine native speakers yielded substantial inter-annotator agreement on the main tasks, with clear language effects: Italian shows the strongest PRCT signal, while Portuguese has higher hyperpartisan prevalence. As a benchmark, \textsc{PartisanLens} shows that even light supervision (few-shot) improves over zero-shot and model scale helps, but can be matched by well-chosen demonstrations.  Moreover, the three main categories are separated in difficulty, with PRCT showing the easiest to detect, hyperpartisan next, and stance remaining the most challenging.

We also evaluated LLMs as annotators. Strong language models such as \texttt{Llama~3.3--70B} remain below human agreement across tasks and languages. Conditioning LLMs on realistic annotator profiles led them to better reflect human-like variability, improving alignment with individual annotation patterns for PRCT across languages, showing mixed effects for hyperpartisan content, and leaving stance differences largely unchanged. We release data, code and all materials to support reproducibility and future work on multilingual partisan narratives in European media. 

Future work will $(i)$ enlarge language coverage to other European languages, $(ii)$ add document-level context to enable discourse-aware models, and $(iii)$ conduct cross-lingual transfer experiments to better isolate language and domain effects.

\section*{Computational Resources}

Experiments were conducted using a private infrastructure, which has a carbon efficiency of 0.432 kgCO2eq/kWh. A cumulative of 21 hours of computation was performed on hardware of type RTX A6000 (TDP of 300W). Total emissions are estimated to be 1.94 kgCO2eq of which 0 per cent were directly offset. Estimations were conducted using MachineLearning Impact calculator \cite{lacoste2019quantifying}.

\section*{Limitations}

Our dataset is necessarily small: constructing a fine-grained, multi-task resource required substantial expert effort (approximately 66 hours of annotation by nine native speakers, experts in the matter). Because Media Cloud provides headlines but not article bodies, we restricted the collection to headlines to avoid copyright issues around full-text scraping; as a result, models and analyses operate without broader article context. The set of rhetorical biases is also limited by design. Following \citet{Maggini}, we focused on the phenomena most influential in shaping hyperpartisan style (loaded language, appeal to fear, name-calling). Expanding to additional techniques would have significantly increased annotator training time and burden, which we judged inappropriate for a volunteer cohort.

Cultural and contextual sensitivity is another constraint. Italian, Spanish, and Portuguese political discourses are embedded in specific historical and social frames that LLMs may not fully capture. That is, what appears hyperpartisan to a model may be mischaracterised relative to local norms. Finally, we acknowledge that labelling content as ``hyperpartisan'' contains normative and political judgments that extend beyond purely linguistic analysis. Our goal is analytical clarity and reproducibility, but these categories should be interpreted with care and with awareness of their social implications. However, we caution that the dataset could be misused by malicious actors to train models capable of generating hyperpartisan or discriminatory content.

\section*{Ethical Statement}

All annotations were performed by volunteers who participated freely, without time pressure or coercion. Annotators could pause or withdraw at any point, and they received compensation for their work commensurate with the expected effort. Socio-economic and political self-reports used to derive personas were fully anonymised. We collected only coarse, non-identifying attributes (e.g., age range, self-reported gender) and ideology scores that were standardised before use. Raw responses and any potentially identifying information were not retained or released; persona prompts were generated from these de-identified summaries, and the public release includes only anonymised labels and headlines.

\section*{Acknowledgments}
The authors thank the funding from the Horizon Europe research and innovation programme under the Marie Skłodowska-Curie Grant Agreement No. 101073351. Views and opinions expressed are however those of the author(s) only and do not necessarily reflect those of the European Union or the European Research Executive Agency (REA). Neither the European Union nor the granting authority can be held responsible for them. The authors thank the financial support supplied by the grant PID2022-137061OB-C21 funded by MI-CIU/AEI/10.13039/501100011033 and by “ERDF/EU”. The authors also thank the funding supplied by the Consellería de Cultura, Educación, Formación Profesional e Universidades (accreditations ED431G 2023/01 and ED431C 2025/49) and the European Regional Development Fund, which acknowledges the CITIC, as a center accredited for excellence within the Galician University System and a member of the CIGUS Network, receives subsidies from the Department of Education, Science, Universities, and Vocational Training of the Xunta de Galicia. Additionally, it is co-financed by the EU through the FEDER Galicia 2021-27 operational program (Ref. ED431G 2023/01).

\bibliography{ARR/ARR}


\appendix

\section{Data Statement}
We report the Data Statement using the schema proposed by \citet{bender2018data}.

\paragraph{Curation rationale} The selected texts consist of news headlines collected via API from the platform Media Cloud, an open-source media research project, enabling the study of news and information flow globally. The goal in selecting texts is to cover and analyse the spread, the linguistic patterns and the stance of headlines on European Immigration, detecting the hyperpartisan and neutral content.

\paragraph{Language variety} Referring to the language tags contained in BCP-47, we list the language variety in our dataset: it-IT, es-ES, pt-BR, pt-PT.

\paragraph{Annotator demographic}

Table \ref{tab:annotator-demographic} presents the demographics of the annotators who participated in the curation of this dataset.

\begin{table*}[ht]
    \centering
    \small
    \begin{tabular}{lcccccc}
    \toprule
     & Age & Gender & Race/Ethnicity & Native language & Socioeconomic status & Training \\
    \midrule
    % --- SPA ---
    SPA 1 & 27 & Male   & Hispanic    & Spanish    & employed      & STEM (PhD) \\
    SPA 2 & 28 & Female & Caucasian   & Spanish    & PhD student   & STEM (MSc) \\
    SPA 3 & 28 & Female & White       & Galician   & employed      & Humanities (MA) \\
    % --- ITA ---
    ITA 1 & 28 & Male   & Italian     & Italian    & PhD student   & Humanities (PhD) \\
    ITA 2 & 27 & Male   & Caucasian   & Italian    & PhD student   & Humanities (PhD) \\
    ITA 3 & 26 & Female & White       & Italian    & unemployed    & Humanities (MA) \\
    % --- PT ---
    PT 1  & 55 & Female & Caucasian   & Portuguese & employed, married & Humanities (MA) \\
    PT 2  & 26 & Female & Caucasian   & Portuguese & Student       & Humanities (BA) \\
    PT 3  & 24 & Female & White       & Portuguese & employed      & STEM (MSc) \\
    \bottomrule
    \end{tabular}
    \caption{Participant demographic overview}
    \label{tab:annotator-demographic}
\end{table*}

\paragraph{Text characteristics} The dataset is composed of news headlines on European Immigration collected from a wide range of media outlets covering the whole political leaning spectrum. For this reason, the texts are short. Due to the nature of the task and the topic, some headlines may refer to immigration using racist and toxic language, spreading xenophobic messages, which could result in the adoption of a conspiracist tone with specific linguistic and topic patterns (PRCT). The primary goal of the dataset is the detection of extremely polarised news to understand how they address these themes.

\section{Sociodemographic Data}

Table \ref{tab:sociodemodata} illustrates the sociodemographic data we collected for each annotator. All of those dimensions have been used to craft the Persona Prompts.

\begin{table}[ht]
    \centering
    \small
    \begin{tabular}{p{4.3cm}p{2.5cm}}
        \toprule
        \textbf{Attribute} & \textbf{Value} \\
        \midrule
        Gender: & Male \\
        Education: & Doctorate \\
        Background: & Humanities \\
        Age: & 28 \\
        COMPASS Economic: & -5.25 \\
        COMPASS Social: & -6.21 \\
        PRISM Government: & 100 Technocracy \\
        PRISM Economy: & 84 Mixed Markets, 16 Ordoliberalism \\
        PRISM Society: & 100 Freedom \\
        Native Language: & Italian \\
        Native or Migrant: & Migrant \\
        Ethnicity: & Italian \\
        Employment Status: & Student \\
        Marital Status: & Partnered \\
        Religion: & Atheist \\
        Disability Status: & No \\
        \bottomrule
    \end{tabular}
    \caption{Example of sociodemographic data. \textit{COMPASS Economic} ranges from negative to positive values, where lower scores indicate a left-leaning economic stance, and higher scores indicate a right-leaning stance. \textit{COMPASS Social} follows the same logic, with lower values reflecting a libertarian orientation and higher values an authoritarian one.}
    \label{tab:sociodemodata}
\end{table}

\section{Annotation Tool}

\begin{figure*}[t]
  \centering
  \includegraphics[width=0.9\textwidth]{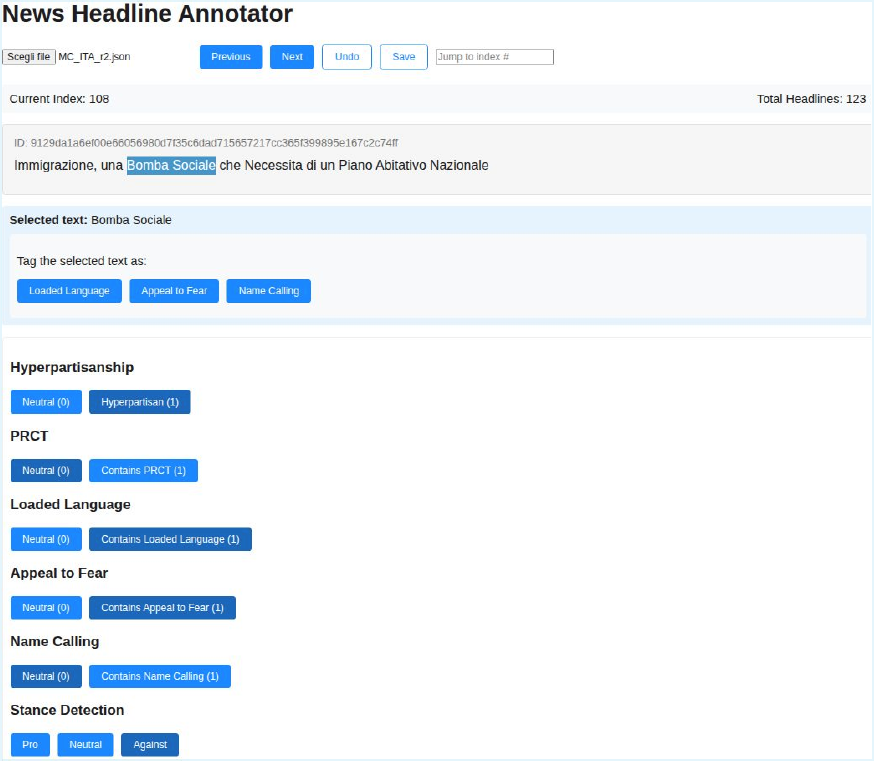}
  \caption{Example of the custom HTML platform used for the annotation process.}
  \label{fig:platform_example}
\end{figure*}

The custom platform allows annotators to annotate one headline at a time using the labels introduced in \ref{sec:DatasetCollection}. The output file is a csv file.

\section{Prompts}

Tables \ref{tab:zero-shot-prompt} and Table \ref{tab:zero-shot-prompt-persona} show the prompts used in the experiments.

\begin{table}[h!]
    \scriptsize
    \tabcolsep=0.08cm
    \centering
    \begin{tabular}{p{7cm}}
    \toprule
    You are a strict JSON generator. Analyze the following news headline and output a JSON object with this exact format:
    \begin{lstlisting}[language=json]
    {
      "hyperpartisan": "<Boolean>",
      "prct": "<Boolean>",
      "stance": "<pro|against|neutral>",
    }
    \end{lstlisting}
    \textbf{Hyperpartisan}: Does it show strong ideological bias or use emotionally charged language? (True/False) \\
    \textbf{PRCT}: Does it contain Population Replacement Conspiracy Theory content claiming a deliberate plan to replace native populations? (True/False) \\
    \textbf{Stance}: What is its stance toward immigration policies? (pro/against/neutral).\\
    DO NOT include any commentary or explanation. Only return valid JSON. \\
    Headline: <headline> \\
    \bottomrule
    \end{tabular}
    \caption{Zero-shot and few-shot prompt.}
    \label{tab:zero-shot-prompt}
    \end{table}
\normalsize

\begin{table}[h!]
    \scriptsize
    \tabcolsep=0.08cm
    \centering
    \begin{tabular}{p{7cm}}
    \toprule
    
    You are a \textit{<gender>} individual with a \textit{<age>} years old, holding a master's degree in \textit{<subject>}. Your COMPASS test values indicate you are leaning towards \textit{<political leaning>}. 
    
    You advocate for \textit{<PRISM\_Government>}, support \textit{<PRISM\_Economic>} matters, and align with \textit{<PRISM\_Society>}. 
    
    Your native language is \textit{<language>}, and you identify as \textit{<religion>}.
    
    You are a strict JSON generator. Analyze the following news headline and output a JSON object with this exact format:
    \begin{lstlisting}[language=json]
    {
      "hyperpartisan": "<Boolean>",
      "prct": "<Boolean>",
      "stance": "<pro|against|neutral>",
    }
    \end{lstlisting}
    \textbf{Hyperpartisan}: Does it show strong ideological bias or use emotionally charged language? (True/False) \\
    \textbf{PRCT}: Does it contain Population Replacement Conspiracy Theory content claiming a deliberate plan to replace native populations? (True/False) \\
    \textbf{Stance}: What is its stance toward immigration policies? (pro/against/neutral).\\
    DO NOT include any commentary or explanation. Only return valid JSON. \\
    Headline: <headline> \\
    \bottomrule
    \end{tabular}
    \caption{Zero-shot Persona prompts.}
    \label{tab:zero-shot-prompt-persona}
    \end{table}
\normalsize

\section{Determinantal Point Process}\label{sec:APP_DPP}
Determinantal Point Processes (DPP) are probability distributions over sets of points, used in physics, statistics, and machine learning for selecting diverse and representative subsets. 

Given a large set: 
\[
A = \{1, 2, \dots, N\}
\]
with items 
\[
I_A = \{x_1, x_2, \dots, x_N\},
\]
subset selection is computationally expensive since it involves evaluating \(2^M\) possibilities. DPP addresses this by first embedding data points (here, with Sentence-BERT), then constructing a kernel matrix:
\[
K_{ij} = k(x_i, x_j)
\]
based on similarity. The probability of selecting a subset \(Y \subset A\) is:
\[
P(Y) = \frac{\det(K_Y)}{\det(K + I)},
\]
where \(K_Y\) is the restriction of \(K\) to the indices in \(Y\).

The goal is to select the subset:
\[
Y_{\text{best}} = \arg\max_{Y \subset A, \, |Y|=k} \det(L_Y),
\]
of fixed size \(k\), which maximises diversity in the representation space. Overall, DPP sampling ensures that selected datapoints are both representative and diverse.

\section{Classification Errors}

Across models and prompting strategies, we observe consistent disparities in how negative and positive instances are classified. All systems show a strong bias toward predicting the negative class, achieving high true-negative rates but struggling to detect positives reliably. Zero-shot Llama 8B exemplifies this imbalance, with very few positive predictions despite strong performance on negatives. Few-shot prompting generally improves recall for the positive class, though often at the cost of misclassifying more negatives. Fine-tuning mitigates this trade-off for some models, particularly Llama 70B, leading to more balanced predictions and a notable increase in positive recall without severe degradation of negative precision. Rationales provide mixed results: while they can improve reasoning transparency, their effect on classification accuracy is unstable and model-dependent. Nemo models remain highly conservative, showing the strongest skew toward negative predictions in all configurations. Overall, the results highlight the difficulty of achieving balanced detection in low-resource or imbalanced settings, emphasising the importance of both fine-tuning and prompting strategies to counteract default negative bias.

\section{Per-language Analysis of Classification Results}
\label{app:lang_analysis}

\begin{table}[h]
\centering
\small
\setlength{\tabcolsep}{5pt}
\resizebox{\columnwidth}{!}{%
    \begin{tabular}{ll rrr rrr rrr}
    \toprule
    \multirow{2}{*}{\textbf{Model}} & \multirow{2}{*}{\textbf{Lang.}} &
    \multicolumn{3}{c}{\textbf{Hyperpartisan}} &
    \multicolumn{3}{c}{\textbf{PRCT}} &
    \multicolumn{3}{c}{\textbf{Stance}} \\
    \cmidrule(lr){3-5}\cmidrule(lr){6-8}\cmidrule(lr){9-11}
    & & P & R & F1 & P & R & F1 & P & R & F1 \\
    \midrule
    \multirow{3}{*}{\texttt{Llama 8B}} 
     & SPA   & 0.75 & 0.75 & 0.75 & 0.68 & 0.64 & 0.66 & \textbf{0.65} & \textbf{0.71} & 0.53 \\      
     & ITA   & 0.72 & 0.63 & 0.62 & 0.84 & 0.73 & 0.75 & 0.57 & 0.44 & 0.45 \\
     & PT   & 0.66 & 0.65 & 0.65 & 0.59 & 0.73 & 0.63 & 0.57 & 0.47 & 0.42 \\  
     \midrule
    \multirow{3}{*}{\texttt{Llama 70B}} 
     & SPA   & 0.67 & 0.68 & 0.65 & 0.67 & 0.82 & 0.71 & 0.46 & 0.53 & 0.43 \\      
     & ITA   & 0.78 & \textbf{0.77} & \textbf{0.77} & 0.85 & \textbf{0.88} & \textbf{0.86} & 0.54 & 0.50 & 0.50 \\
     & PT   & 0.72 & 0.66 & 0.60 & 0.57 & 0.72 & 0.59 & \textbf{0.65} & 0.62 & \textbf{0.62} \\  
     \midrule
    \multirow{3}{*}{\texttt{Nemo}} 
     & SPA   & 0.75 & 0.76 & 0.75 & 0.97 & 0.79 & 0.85 & 0.43 & 0.47 & 0.34 \\      
     & ITA   & \textbf{0.81} & 0.75 & 0.75 & 0.83 & 0.68 & 0.70 & 0.60 & 0.44 & 0.44 \\
     & PT   & 0.61 & 0.61 & 0.60 & \textbf{0.99} & 0.75 & 0.83 & 0.53 & 0.45 & 0.41 \\  
     \bottomrule
    \end{tabular}
}
\caption{Fine-tuning with rationales results by language (macro-averaged).}
\label{tab:experiments-lang}
\end{table}

Table~\ref{tab:experiments-lang} presents the performance by language across Spanish (SPA), Italian (ITA), and Portuguese (PT) under the fine-tuning with rationales setting. Looking at the results, hyperpartisan is robust in all three languages, with the best scores in Italian (F1\textsubscript{MACRO}$=0.77$, \texttt{Llama 3 70B}) and a tie at the top in Spanish (F1\textsubscript{MACRO}$=0.75$, \texttt{Llama 3 8B} and \texttt{Nemo}). Portuguese is lower overall (best 0.65, \texttt{Llama 3 8B}). PRCT shows more extreme language effects depending on the model: Italian peaks (F1\textsubscript{MACRO}$=0.86$, \texttt{Llama 3 70B}), while \texttt{Nemo} is strongest in Spanish (0.85) and Portuguese (0.83), but lower recall (R$\approx$0.79/0.75), indicating conservative detection that favours salient conspiracy markers.

Stance remains the most challenging and language-sensitive task: Portuguese sees the best result (F1\textsubscript{MACRO}$=0.62$, \texttt{Llama~70B}), Italian is moderate (0.50, \texttt{Llama~70B}), and Spanish lags (best F1\textsubscript{MACRO}$=0.53$, \texttt{Llama~8B}). Overall, Italian headlines favour higher PRCT and hyperpartisan detectability (consistent with a stronger signal), whereas Spanish stance is notably difficult, likely reflecting a skew toward neutral headlines and subtler polarity cues. Regarding Portuguese, it benefits most for stance from large-scale fine-tuning.

\end{document}